\documentclass[letterpaper,10pt,conference]{ieeeconf}

\IEEEoverridecommandlockouts
\usepackage{cite}
\usepackage{amsmath,amssymb,amsfonts}
\usepackage{graphicx}
\usepackage{xspace}
\usepackage{algorithm}
\usepackage{algpseudocode}
\usepackage{array}
\usepackage{booktabs}
\usepackage{multirow}
\usepackage{wrapfig}
\usepackage{xcolor}
\usepackage{hyperref}
\usepackage{microtype}

\hypersetup{
  colorlinks=true,
  linkcolor=blue,
  citecolor=green,
  urlcolor=cyan
}

\newcommand{\sysname}{\textit{VLA-Stem}\xspace}

\title{VLA-Stem: A Lightweight Safety Mechanism for Vision-Language-Action Models via Reinforcement Learning}

\title{VLA-Stem: A Lightweight Safety Mechanism
for Vision-Language-Action Models via
Reinforcement Learning}

\author{
Bojun Zou$^{1,2*}$,
Wenhao Yu$^{1*}$,
Jiaohao Li$^{1}$,
Jiaxuan Lin$^{1}$,
Guanting Ye$^{3}$,\\
Haohan Geng$^{1}$,
Weihao Yu$^{2}$,
Yanyong Zhang$^{1}$,
Jianmin Ji$^{1\dagger}$\\[5pt]
{\small $^{1}$University of Science and Technology of China (USTC)}\\
{\small $^{2}$Peking University (PKU)}\\
{\small $^{3}$University of Macau (UM)}\\[3pt]
{\footnotesize $^{*}$Equal contribution.
\quad $^{\dagger}$Corresponding author.}
}

\begin{document}
\maketitle
\thispagestyle{empty}
\pagestyle{empty}

\begin{abstract}
Vision-Language-Action (VLA) models have emerged as a prominent paradigm for end-to-end robotic control, integrating high-level reasoning with low-level action execution. However, most existing VLA models are primarily designed for static manipulation tasks in unconstrained environments, failing to account for safety-critical scenarios that involve spatial constraints or dynamic obstacles.
Fine-tuning VLAs on obstacle-avoidance data is a straightforward solution, but collecting such data is costly, especially for dynamic scenarios, and may degrade their original task capabilities.
To address these limitations, we propose \sysname, a lightweight, reinforcement-learning-based safety mechanism for VLA systems. 
Without fine-tuning the underlying VLA policy, \sysname converts its action chunks into motion references and uses trajectory-level guidance to preserve their spatial intent while allowing flexible motion and timing adjustments.
Conditioned on these references, robot proprioception, and local obstacle information, the Stem policy adaptively modifies execution at inference time, preserving the original task intent while enabling responsive adaptation to spatial constraints and dynamic obstacles.
To achieve robust and sample-efficient policy learning in complex environments, we adopt a teacher--student distillation framework combined with curriculum learning. We evaluate \sysname on reference following, static spatial constraints, and dynamic obstacle avoidance. Across six constrained simulation settings, \sysname achieves 52.8\% average collision-free task success versus 23.7\% for the strongest safety baseline, while preserving base-policy performance in obstacle-free settings; real-world experiments further show consistent collision reductions under static and dynamic obstacles. 
These results demonstrate that \sysname couples high-level VLA task intent with rapid, reflex-like responses to immediate physical risks, providing a brainstem-inspired mechanism for robust robot execution.
\end{abstract}

\section{Introduction}
Embodied intelligence aims to integrate robotics with artificial intelligence for general real-world interaction, with Vision-Language-Action (VLA) models emerging as a prominent direction. State-of-the-art VLA models~\cite{pi0.5} adopt end-to-end architectures that combine high-level capabilities with low-level action execution. This resembles human behavior, where deliberate reasoning guides bodily actions to accomplish diverse real-world tasks.
However, we argue that this paradigm overlooks another fundamental aspect of biological intelligence: basic physiological and sensorimotor functions, such as postural regulation and rapid reflexive responses, which operate through lower-level neural circuits including the brainstem.
In contrast, most existing VLA policies lack an explicit counterpart to such lower-level responsive regulation.
Therefore, VLA policies should not only learn what to do and how to do it, but also remain responsive to the safety demands of physical interaction.



\begin{figure}[t]
    \centering
    \includegraphics[width=\columnwidth]{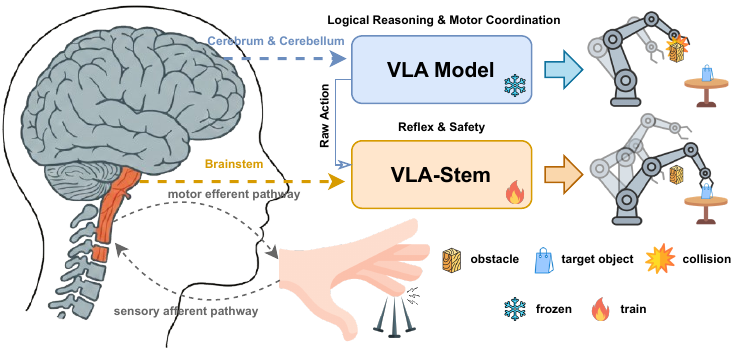}
    \caption{This figure illustrates the biological motivation behind our method. The left side depicts the interaction between higher-level motor control and lower-level sensorimotor regulation: task-directed motor commands are continuously shaped by rapid responses to immediate physical conditions. In this functional analogy, higher-level task control and motor coordination correspond to the base VLA, while the brainstem-inspired lower-level regulation corresponds to our proposed \sysname. The right side shows how \sysname adapts the VLA-directed motion to bypass an obstacle while continuing toward the target object.}
    \label{fig:first}
\end{figure}


Building on this perspective, we aim to endow VLA systems with the ability to react to immediate physical risks while remaining coordinated with their task-directed behavior (Fig.~\ref{fig:first}). Existing approaches mainly follow two directions. One is to learn collision-aware manipulation behaviors from large-scale avoidance data~\cite{yang2025deep}; however, collecting such data is labor-intensive and costly, especially in dynamic environments, and adapting a pretrained VLA requires additional training of the task policy itself. Another direction is to augment a frozen VLA with a conventional safety module. For example, VLSA~\cite{hu2025vlsa} introduces control-theoretic constraints to modify unsafe VLA actions during execution. While this modular design provides an explicit mechanism for satisfying local geometric constraints, it primarily focuses on action-space safety and ignores the coordination between safety responses and the task-directed behavior of the VLA, leading to restricted corrective behaviors, reduced task success, and limited adaptability in complex dynamic environments. This motivates a mechanism that does not treat safety adaptation independently of VLA task intent, but instead tightly integrates safety responses with VLA task execution.

To address these limitations, we propose \sysname, a lightweight, brainstem-inspired RL-based safety mechanism that couples VLA task intent with rapid responses to physical risks. Specifically, \sysname uses the short-horizon trajectory generated by the base VLA as a reference for its intended motion, and jointly conditions on this reference trajectory, robot proprioception, and local obstacle information to generate executable actions. To support such flexible execution, we introduce a trajectory-level guidance objective that preserves the spatial intent of the motion reference while relaxing strict temporal alignment. We further adopt teacher--student distillation with curriculum learning to facilitate structured exploration and improve the learned policy's recovery capability under safety-induced deviations. As a result, \sysname improves execution safety and robustness without sacrificing task performance or incurring additional training cost for the base policy.

We evaluate \sysname across multiple manipulation tasks and scenarios, including obstacle-free execution, static spatial constraints, and dynamic obstacle avoidance. Results show that \sysname reduces safety violations while maintaining base-policy task performance, demonstrating its potential as an inference-time safety mechanism for VLA policies.
Our main contributions are summarized as follows:
\begin{itemize}
\item We propose \sysname, a lightweight, reinforcement-learning-based safety mechanism for VLA models that directly integrates VLA task intent with runtime safety adaptation to physical constraints, without fine-tuning the underlying policy.
\item We design a trajectory-level guidance objective that uses the base VLA trajectory as a reference motion, allowing flexible local motion and timing adjustments while preserving task behavior. We further introduce a teacher--student distillation framework with curriculum learning to improve recovery from obstacle-avoidance-induced deviations.
\item We conduct extensive experiments across multiple manipulation scenarios in both simulation and the real world,  demonstrating improved collision-free task success and direct sim-to-real transfer of the Stem policy without real-world training or fine-tuning.
\end{itemize}

\section{Related Work}
\subsection{Vision-Language-Action Models}
Vision-Language-Action (VLA) models have emerged as an end-to-end paradigm for robot manipulation, conditioning action generation on visual observations and language instructions\cite{ye2026c,yip2026sope}.
Representative systems, including OpenVLA~\cite{kim2024openvla}, RT-X~\cite{zitkovich2023rt,brohan2022rt}, RDT~\cite{liu2024rdt}, $\pi_0$~\cite{black2024pi_0}, and $\pi_{0.5}$~\cite{pi0.5}, have continued to improve the scale, generalization, and action-generation capability of robot policies ~\cite{xiong2026cac, geng2026trajwam}.

Although VLA models demonstrate strong generalization, they remain constrained by high-quality data requirements and limited robustness in out-of-distribution scenarios.
Recent work has explored reinforcement learning for VLA adaptation, such as VLA-RL~\cite{lu2025vlarl} and $\pi^*_{0.6}$~\cite{amin2025pi}. 
While these methods improve task-level robustness, they still primarily optimize policies for task completion and adaptation, without explicitly treating execution safety in challenging manipulation scenarios as a central objective.


\subsection{Safety for Manipulators}
Safe trajectory generation has been widely studied through sampling-based planners such as RRT~\cite{lavalle2001rapidly} and PRM~\cite{kavraki1996probabilistic}, optimization-based methods such as CHOMP~\cite{ratliff2009chomp}, TrajOpt~\cite{TrajOpt}, and STOMP~\cite{kalakrishnan2011stomp}, as well as MPC. Although widely used, these methods often require substantial manual effort due to their complexity and the need for extensive retuning across different scenarios.
Learning-based methods~\cite{yu2024ldp}, including Motion Planning Diffusion~\cite{carvalho2023motionplanningdiffusion}, Deep Reactive Policy~\cite{yang2025deep}, and URPlanner~\cite{ying2025urplanner}, improve efficiency and generalization through demonstrations or reinforcement learning~\cite{yu2024pathrl}. 
However, they typically act as standalone trajectory generators and are difficult to directly integrate with VLA policies without disrupting the task-directed behavior of the base policy.

Beyond trajectory generation, recent work has also explored safety mechanisms for end-to-end policies. 
CBF-based safe RL methods~\cite{saferl1,saferl2} enforce safety through constrained policy optimization, with SafeVLA~\cite{zhang2026safevla} extending this direction to VLA policies, while safety filters such as RAIL~\cite{jung2025rail} augment frozen policies with external safety modules. 
However, filtering-based corrections can become less reliable in complex or dynamic environments and may compromise base-policy performance on challenging tasks.

\subsection{Test-Time VLA}
Test-time VLA~\cite{deng2025survey} is a recent paradigm that adjusts VLA actions during deployment, avoiding costly fine-tuning.
VLA-Reasoner~\cite{guo2025vlareasoner} improves actions through online MCTS, while V-GPS~\cite{nakamoto2024steering} selects candidate actions using a pretrained value function. 
However, these methods evaluate multiple candidate action sequences at inference time, boosting performance at the expense of real-time efficiency. In contrast, our method performs lightweight VLA-conditioned execution with a compact Stem policy, enabling efficient safety augmentation without candidate search or VLA fine-tuning.
\section{Method}

\begin{figure*}[t]
    \centering
    \includegraphics[width=0.95\textwidth]{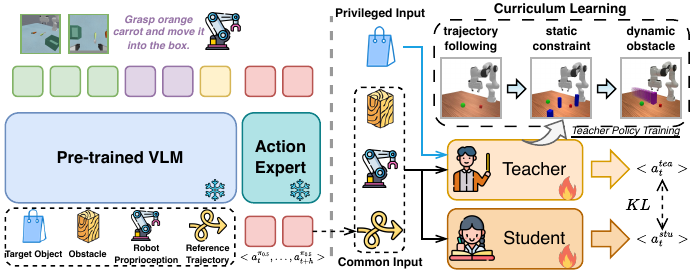}
    \caption{\sysname pipeline. The teacher is trained using curriculum learning, progressing from reference trajectory following to static obstacle avoidance and then to dynamic obstacle avoidance, with gradually increasing difficulty. The teacher policy takes as input the trajectory sequence produced by the base VLA policy, robot proprioception, and obstacle information, and additionally leverages privileged task information about the target object to accelerate training. Finally, the teacher policy is used to distill a student policy, removing the dependence on task-specific privileged information at deployment.}
    \label{fig:overview}
\end{figure*}

An overview of the proposed framework is illustrated in Fig.~\ref{fig:overview}. 
We first formulate VLA-conditioned execution, where the frozen VLA provides a task-directed motion reference that conditions the Stem policy (\S\ref{sec:vla_conditioned_execution}).
We then introduce trajectory-level guidance to preserve the spatial intent of this reference while allowing flexible motion and timing adjustments (\S\ref{sec:trajectory_guidance}).
Next, we present teacher--student distillation for learning recovery from execution deviations using privileged task information during training (\S\ref{sec:distillation}).
Finally, we describe the RL objective and curriculum that progressively train the Stem policy from trajectory following to static and dynamic obstacle interactions (\S\ref{sec:curriculum}).

\subsection{VLA-Conditioned Execution}
\label{sec:vla_conditioned_execution}
We formulate \sysname as a coupled execution architecture in which the frozen VLA provides a short-horizon task-directed motion reference, while the Stem policy conditions jointly on this reference and the current physical context to generate executable actions.


Execution is organized into intervals of $h$ control steps, each associated with one VLA action chunk.
Let $c$ denote such an execution interval and $t_c$ its first control step.
At $t_c$, the frozen VLA policy $\pi_{\mathrm{VLA}}$ is queried with its observation $\mathbf{o}_{t_c}^{\mathrm{VLA}}$ and produces an $h$-step action chunk
\begin{equation}
    \mathcal{A}_c^{\mathrm{VLA}}
    =
    \pi_{\mathrm{VLA}}
    \left(
        \mathbf{o}_{t_c}^{\mathrm{VLA}}
    \right).
\end{equation}
Rather than being executed directly, this action chunk is converted into a motion reference that conditions Stem execution over the current interval.

Let $T_t^{w} \in SE(3)$ denote the end-effector pose in the world frame.
We map the VLA action chunk to the corresponding future end-effector poses $\{T_{t_c+1}^{w}, \dots, T_{t_c+h}^{w}\}$ and express them relative to the end-effector pose $T_{t_c}^{w}$ at the beginning of the interval:
\begin{equation}
    \Delta T_{c,k}
    =
    \left(T_{t_c}^{w}\right)^{-1} T_{t_c+k}^{w},
    \qquad
    \tau_c^{\mathrm{rel}}
    =
    \left\{
        \Delta T_{c,k}
    \right\}_{k=1}^{h}.
\end{equation}
The resulting trajectory $\tau_c^{\mathrm{rel}}$ serves as the task-directed motion reference until the next VLA query.

At each control step $t$ within the interval, the Stem policy also receives the current physical context, consisting of robot proprioception and local obstacle information.
The proprioceptive state $\mathbf{x}_t \in \mathbb{R}^{n}$ contains the joint positions and end-effector pose.
For each obstacle $j$, the local obstacle information consists of its center
$\mathbf{c}_{j,t}^{w} \in \mathbb{R}^{3}$ in the world frame and its bounding-box dimensions $\mathbf{d}_{j,t} = (d_x, d_y, d_z)$.
We represent the set of local obstacles as
\begin{equation}
    \mathbf{o}_t^{\mathrm{obs}}
    =
    \left\{
        \left(
            \mathbf{c}_{j,t}^{w},
            \mathbf{d}_{j,t}
        \right)
    \right\}_{j=1}^{M},
\end{equation}
where $M$ denotes the number of local obstacles.

Conditioned on the active VLA motion reference and the current physical context, the Stem policy generates an executable action at each control step:
\begin{equation}
    \label{eq:coupled_execution}
    a_t^{\mathrm{exec}}
    =
    \pi_{\mathrm{stem}}
    \left(
        \mathbf{x}_t,\,
        \mathbf{o}_t^{\mathrm{obs}},\,
        \tau_c^{\mathrm{rel}}
    \right),
    \qquad
    t_c \leq t < t_c+h.
\end{equation}

After these $h$ execution steps, the VLA is queried again with the updated observation to generate the action chunk for the next execution interval.

\subsection{Trajectory-Level Guidance for Spatial Intent Preservation}
\label{sec:trajectory_guidance}
Many runtime safety mechanisms operate by correcting the current nominal action at each control step. 
Such local correction can satisfy immediate physical constraints but does not account for how the corrected robot state affects the remaining VLA action sequence.
The same issue arises when step-wise action or waypoint alignment is directly used to guide RL exploration and sampling.
The resulting robot state may therefore become misaligned with subsequent time-indexed actions or waypoints along the reference trajectory.
Repeated step-wise correction or alignment can therefore produce discontinuous corrections and accumulate deviation from the VLA-directed motion.
This accumulated mismatch can further shift closed-loop execution away from the state distribution expected by the base VLA.

To address this issue, we introduce trajectory-level guidance.
Instead of aligning execution with a fixed action or waypoint at each time step, we measure its deviation from local segments of the VLA reference trajectory.
This preserves spatial consistency with the task-directed motion at the trajectory level.
We further preserve forward progression while relaxing strict temporal correspondence.

Specifically, for reward computation, we use the world-frame form of the same active reference trajectory,
$\tau_c^{\mathrm{ref}}
=\{(\mathbf{p}_i^{\mathrm{ref}},
\mathbf{q}_i^{\mathrm{ref}})\}_{i=1}^{h}$.
For each segment between consecutive waypoints, we project
the current end-effector position onto the positional segment
and interpolate its endpoint orientations using normalized
quaternion interpolation with the same projection coefficient.
The segment deviation is
$d_i=d_i^{\mathrm{pos}}+\alpha d_i^{\mathrm{rot}}$,
where $d_i^{\mathrm{pos}}$ is the Euclidean distance to the
projected position, $d_i^{\mathrm{rot}}$ is the rotational
geodesic distance to the interpolated orientation, and
$\alpha$ balances the two terms.

To avoid matching delayed or detoured execution to previously
traversed segments, we maintain a non-decreasing segment index
$m_t$ and restrict matching to a forward corridor:
\begin{equation}
    \mathcal{I}_t^{+}
    =
    \{m_{t-1},\ldots,\min(m_{t-1}+K,h-1)\},
\end{equation}
where $K$ is the look-ahead range.
At the start of each execution interval, we reset $m_{t_c-1}=1$.
The matched segment and trajectory-level deviation are
\begin{equation}
    i_t^\ast = \arg\min_{i\in\mathcal{I}_t^{+}} d_i,
    \qquad
    d_t^\ast = d_{i_t^\ast}.
\end{equation}
We then update the progress index as
$m_t=\max(m_{t-1},i_t^\ast)$.
This constrains reference matching to move forward while
allowing execution to remain near the current segment
during waiting or local detours.

We map the deviation to a dense trajectory-following reward:
\begin{equation}
    r_t^{\mathrm{follow}}
    =
    \exp\!\left(-k_{\mathrm{pose}}d_t^\ast\right),
\end{equation}
where $k_{\mathrm{pose}}$ controls sensitivity to reference
deviations.
This forms the trajectory-following component of
$r_t^{\mathrm{guidance}}$ in Sec.~\ref{sec:curriculum}.
During the early curriculum stages, the auxiliary progress
bonus uses the normalized index advancement
$(m_t-m_{t-1})/(h-1)$.

\subsection{Teacher--Student Distillation}
\label{sec:distillation}


Trajectory-level guidance limits execution drift by maintaining spatial consistency with the VLA reference.
However, some constrained interactions require larger deviations that cannot be resolved through local trajectory adjustment alone.
Such deviations can alter subsequent observations and VLA references, pushing closed-loop execution away from the state distribution expected by the base VLA.
Recovering from such states with reinforcement learning alone is challenging, as the guidance reward is task-agnostic and the deployable observation lacks explicit task-state information, making the recovery direction ambiguous.
To address this challenge, we adopt a teacher--student distillation framework.
For the teacher, we augment the deployable observation with privileged task-level information $o_{\text{extra}}$; for example, in pick-and-place tasks, the positions of the target object are provided as $o_{\text{extra}}$. 
The privileged information is introduced only as an additional teacher observation during training and does not modify the reward function.
Thus, the teacher and student are optimized under the same reward design, while differing in their observation spaces. 
These privileged inputs provide direct task information during training, making exploration more structured and helping the teacher learn recovery behaviors more efficiently. 
Since such privileged information is not assumed to be available at deployment time, we distill the teacher policy into a student policy that relies only on the deployable observation space without privileged inputs. Specifically, we optimize the student policy using a loss function that combines reinforcement learning with a distillation loss~\cite{ze2025twist}:
\begin{equation}
L(\pi_{\text{stu}}) = L_{\text{RL}}(\pi_{\text{stu}}) + \lambda D_{\text{KL}}(\pi_{\text{stu}} \parallel \pi_{\text{tea}}),
\end{equation}
where $L_{\mathrm{RL}}$ denotes the PPO~\cite{schulman2017proximal} objective under the reward function defined in Sec.~\ref{sec:curriculum}, and $D_{\mathrm{KL}}$ encourages the student policy to imitate the action distribution of the privileged teacher.
We set $\lambda=0.1$ to balance task optimization and teacher imitation.
After distillation, the student policy serves as the final deployable policy of \sysname, retaining the teacher's recovery behavior while removing the dependence on privileged task information.


\subsection{RL Optimization and Curriculum Learning}
\label{sec:curriculum}
We optimize the Stem policy with PPO~\cite{schulman2017proximal} using a reward that jointly captures task completion, collision avoidance, and consistency with the VLA motion reference:
\begin{equation}
r_t
=
w_{\mathrm{task}} r_t^{\mathrm{task}}
+
w_{\mathrm{coll}} r_t^{\mathrm{coll}}
+
w_{\mathrm{guidance}} r_t^{\mathrm{guidance}}.
\end{equation}
Here, $r_t^{\mathrm{task}}$ provides sparse success feedback, with an auxiliary grasp-event bonus to facilitate learning; each event is rewarded only once per episode.
The collision term $r_t^{\mathrm{coll}}$ is $-1$ whenever a collision or spatial-constraint violation occurs and $0$ otherwise.
The guidance term follows the trajectory-level objective in Sec.~\ref{sec:trajectory_guidance} and is augmented during the early curriculum stages with a progress bonus proportional to the normalized forward advancement of the matched reference segment.
Reward weights vary across curriculum stages, while the reward design is shared across all tasks.

To progressively introduce the behaviors required for VLA-conditioned execution, we organize training into three sequential curriculum stages~\cite{bengio2009curriculum}. 
The first stage trains $\pi_{\mathrm{stem}}$ in obstacle-free environments to follow the task-directed motion provided by the frozen VLA. The second stage introduces static spatial constraints, encouraging the policy to make local spatial adjustments while remaining aligned with the reference trajectory. The third stage introduces moving obstacles, requiring the policy to jointly adjust local motion and execution timing, including waiting when necessary before resuming task progression.

\section{Experiment}

\begin{table*}[t]
\centering
\caption{Experimental results on simulation tasks. Higher SUCC is better and lower COLL is better.}
\label{tab:exp}
\renewcommand{\arraystretch}{1.1}
\setlength{\tabcolsep}{4pt}

\resizebox{\textwidth}{!}{
\begin{tabular}{
>{\centering\arraybackslash}p{2.0cm}
>{\centering\arraybackslash}p{1.1cm}
*{9}{>{\centering\arraybackslash}p{1.2cm}} 
}
\toprule

\multirow{2}{*}{\textbf{Method}} &
\multirow{2}{*}{\textbf{Metric}} &
\multicolumn{3}{c}{\textbf{PickCube}} &
\multicolumn{3}{c}{\textbf{Object Stowing}} &
\multicolumn{3}{c}{\textbf{StackCube}} \\
\cmidrule(lr){3-5}\cmidrule(lr){6-8}\cmidrule(lr){9-11}

& & Normal & Static & Dynamic & Normal & Static & Dynamic & Normal & Static & Dynamic \\

\midrule



\multirow{2}{*}{$\pi_{0}$}
& SUCC$\uparrow$ & 0.6884 & 0.0106 & 0.0044 & 0.4322 & 0.0006 & 0.0559 & 0.6506 & 0.1316 & 0.1453 \\
& COLL$\downarrow$ & -- & 0.9422 & 0.9703 & -- & 0.5363 & 0.6809 & -- & 0.6478 & 0.5484 \\

\midrule

\multirow{2}{*}{$\pi_{0.5}$}
& SUCC$\uparrow$ & 0.7166 & 0.0325 & 0.0131 & 0.6138 & 0.0000 & 0.0256 & 0.7450 & 0.1725 & 0.1247 \\
& COLL$\downarrow$ & -- & 0.8781 & 0.9328 & -- & 0.5897 & 0.6156 & -- & 0.7319 & 0.7012 \\

\midrule

\multirow{2}{*}{VLA-finetune}
& SUCC$\uparrow$ & 0.7166 & 0.2847 & 0.1869 & 0.6138 & 0.0072 & 0.4784 & 0.7450 & 0.0062 & 0.0641 \\
& COLL$\downarrow$ & -- & 0.4413 & 0.5156 & -- & 0.2159 & 0.2331 & -- & 0.7469 & 0.6072 \\

\midrule

\multirow{2}{*}{VLSA-EE}
& SUCC$\uparrow$ & 0.7166 & 0.1619 & 0.0959 & 0.6138 & 0.3503 & 0.1678 & 0.7450 & 0.2372 & 0.4078 \\
& COLL$\downarrow$ & -- & 0.2303 & \textbf{0.2847} & -- & 0.2609 & 0.2663 & -- & 0.4509 & 0.2244 \\

\midrule

\multirow{2}{*}{VLSA-ARM}
& SUCC$\uparrow$ & 0.7166 & 0.1516 & 0.0856 & 0.6138 & 0.3506 & 0.1716 & 0.7450 & 0.2334 & 0.4038 \\
& COLL$\downarrow$ & -- & \textbf{0.2253} & 0.2903 & -- & 0.2644 & 0.2578 & -- & 0.3753 & 0.1628 \\

\midrule

\multirow{2}{*}{\sysname}
& SUCC$\uparrow$ & \textbf{0.7325} & \textbf{0.5219} & \textbf{0.4153} & \textbf{0.7061} & \textbf{0.5784} & \textbf{0.5950} & \textbf{0.7521} & \textbf{0.5141} & \textbf{0.5419} \\
& COLL$\downarrow$ & -- & 0.2316 & 0.3122 & -- & \textbf{0.1756} & \textbf{0.1203} & -- & \textbf{0.3353} & \textbf{0.0784} \\

\bottomrule
\end{tabular}}
\end{table*}

\subsection{Experimental Setup}

We conduct simulation experiments in ManiSkill~\cite{taomaniskill3} on three manipulation tasks: PickCube, StackCube, and our Object Stowing task, which requires placing household objects into a target container under spatial constraints.
Across all three tasks, VLA-Stem uses the same Stem architecture, reward design, and training procedure, while keeping the corresponding base VLA frozen throughout Stem training.
Each base VLA is fine-tuned on 100 demonstrations for 30,000 optimization steps, with the resulting checkpoint shared by all compared safety mechanisms for that task.

For each task, we construct three scenarios: normal, static, and dynamic. The normal scenario contains no additional obstacles and evaluates whether \sysname preserves base-policy task performance. The static scenario uses two obstacle placement strategies. The first places obstacles near the target object or placement region to constrain grasping and placement. The second samples points along the nominal object-to-placement path and applies random lateral offsets to generate obstacles around the transfer route. Placements are checked for workspace validity and overlap.

\begin{figure}[t]
    \centering
    \includegraphics[width=\linewidth]{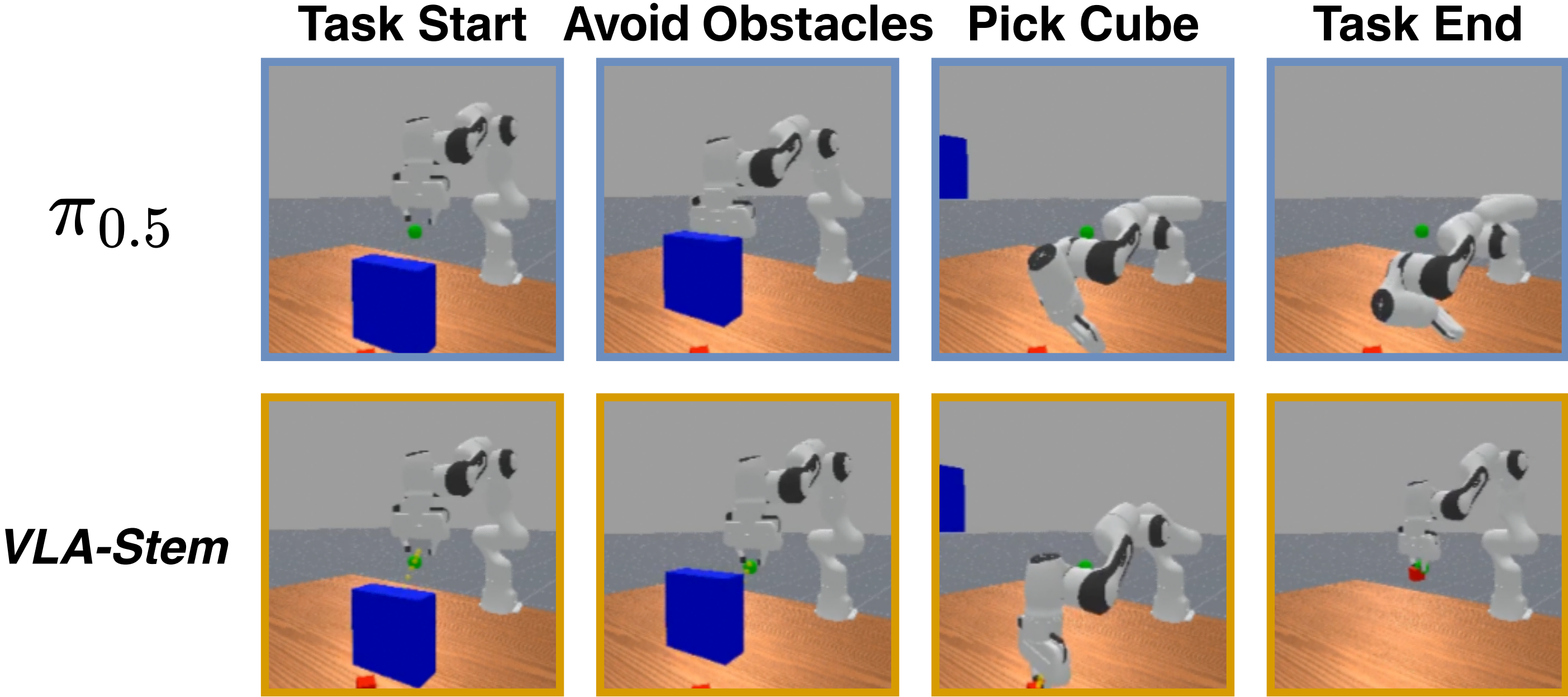}
    \caption{Dynamic obstacle avoidance in simulation.
    Compared with $\pi_{0.5}$, \sysname waits for the moving
    obstacle to pass before resuming task execution.}
    \label{fig:sm_sim}
\end{figure}

In the dynamic scenario, moving obstacles cross the robot's path during either approach or object transfer. We sample a crossing region along the task motion and generate an obstacle trajectory with a randomized starting position, direction, and constant speed. Obstacle crossing times are sampled within a temporal window corresponding to the robot's expected passage through this region, inducing interaction during execution. This setting requires the policy to adapt both its motion and execution timing (Fig.~\ref{fig:sm_sim}).

We compare \sysname with three types of baselines:
(1) the base VLA policies, $\pi_0$ and $\pi_{0.5}$, which reflect the original performance without an additional safety mechanism;
(2) VLA-finetune, which fine-tunes a separate VLA policy for each task and scenario using RRT-based obstacle-avoidance trajectories, with an additional wait primitive in dynamic scenarios. We evaluate VLA-finetune under multiple demonstration budgets and report the best-performing setting in the main results. In our experiments, performance peaks at 1,500 demonstrations, while increasing the data to 2,000 leads to a decline;
and (3) VLSA, a plug-and-play safety baseline using the same $\pi_{0.5}$ checkpoint and ground-truth obstacle information as \sysname.
VLSA applies CBF-based corrections to the nominal end-effector
motion converted from VLA joint actions, leaving gripper commands unchanged.
We evaluate VLSA-EE with end-effector constraints and VLSA-ARM with additional constraints on collision-prone arm links. For each task and scenario, we tune the CBF gain, geometric scales, and optimization weights on validation episodes and report the configuration achieving the highest SUCC.

We report two metrics: (1) SUCC, the proportion of episodes completed without constraint violations or collisions; and (2) COLL, the proportion of episodes with such violations.
For each trained policy, every task--scenario setting is evaluated over 3,200 episodes.

\noindent\textbf{Training details.}
Both teacher and student use MLP actor--critic networks.
For trajectory following, PPO uses 256 parallel environments, a learning rate of $2\times10^{-4}$, 8 update epochs, and a minibatch size of 3,200. For the static and dynamic stages, the corresponding values are 192, $1.5\times10^{-4}$, 12, and 19,200. All stages use a rollout horizon of 100 steps, a discount factor of $\gamma=0.8$, and the Adam optimizer.
Using GPU-parallel simulation, the stage-wise training budgets for the evaluated checkpoints are approximately 38.0M, 23.7M, and 20.6M simulated environment transitions per seed for trajectory following, static, and dynamic training, respectively, including both teacher and student optimization.

\subsection{Results of Simulation Tasks}

\subsubsection{Main Results}
Tab.~\ref{tab:exp}
reports the performance of all methods on the three tasks under the normal, static, and dynamic scenarios.
The results evaluate two key aspects of the coupled execution architecture: preserving the task performance of the base VLA in obstacle-free settings and adapting execution to physical constraints in static and dynamic settings.

\textbf{Task Performance Preservation.}
In the normal scenario, \sysname achieves success rates of 0.7325, 0.7061, and 0.7521 on PickCube, Object Stowing, and StackCube, respectively, compared with 0.7166, 0.6138, and 0.7450 for the corresponding $\pi_{0.5}$ base policies.
Thus, introducing the Stem policy into the execution loop does not reduce the observed task success in any of the three obstacle-free settings.
These results indicate that the coupled execution architecture preserves the task capability of the frozen VLA when safety-induced adaptation is not required.

\textbf{Safety--Task Trade-off.}
Under static and dynamic constraints, the base $\pi_{0.5}$ policy exhibits frequent collisions and correspondingly low collision-free task success.
For example, on PickCube, its SUCC drops sharply from 0.7166 in the normal setting to only 0.0325 and 0.0131 under the static and dynamic obstacles, with collision rates of 0.8781 and 0.9328, respectively.

Increasing obstacle-avoidance supervision through VLA-finetune improves some constrained settings, but its performance remains inconsistent across tasks. Our failure analysis suggests that a single VLA policy has difficulty simultaneously maintaining fine-grained task execution and acquiring reliable safety responses. Even in relatively low-risk cases where no substantial avoidance maneuver is required, the fine-tuned policy can fail on behaviors that the raw VLA performs successfully, particularly during precise manipulation; for example, our failure analysis frequently observes grasp failures on irregular objects such as the carrot in Object Stowing, where the original $\pi_{0.5}$ policy often succeeds. Meanwhile, substantial collision rates remain in several static and dynamic settings despite the additional avoidance supervision. Together, these observations indicate that directly incorporating obstacle-response behavior into the VLA can interfere with its original manipulation behavior without consistently establishing reliable safety adaptation.

VLSA exhibits a different type of limitation.
Its explicit safety optimization can substantially reduce collision rates, but in practice, lower collision rates do not necessarily translate into successful task execution.
On PickCube-Static, for example, VLSA-ARM reduces COLL to 0.2253, slightly below the 0.2316 of \sysname, yet achieves only 0.1516 SUCC compared with 0.5219 for \sysname.
To better understand the gap between collision reduction and task success, we inspect the collision-free failures of VLSA. In static scenes, local corrections near fine-grained phases such as grasping often disrupt the continuity of the VLA motion and lead to grasp failures; when large obstacles block the transfer path, geometric avoidance can instead push execution away from the VLA's nominal motion distribution, making subsequent task recovery difficult. In dynamic scenes, even mild obstacle-induced corrections during early fine-grained phases can disturb the original grasping motion and cause the task to fail before the robot reaches later, more challenging obstacle interactions. Consequently, part of the relatively low collision rate in dynamic settings comes from trajectories that terminate in early task failures, rather than from successful negotiation of the subsequent moving obstacle. These observations further show that local geometric safety correction alone does not necessarily preserve the continuity of VLA-directed task execution.

In contrast, \sysname achieves the highest collision-free task success across the evaluated constrained settings while maintaining competitive collision rates.
On StackCube, it reaches SUCC of 0.5141 and 0.5419 in the static and dynamic settings, with COLL of 0.3353 and 0.0784, respectively. 
Together, these results support the central design of \sysname: using the VLA trajectory as a task-directed motion reference while adapting execution to immediate physical constraints provides a stronger coupling between task intent and responsive safety behavior.


\begin{figure}[t]
    \centering
    \makebox[\linewidth][c]{%
        \begin{minipage}[c]{0.49\linewidth}
            \centering
            \includegraphics[width=\linewidth]{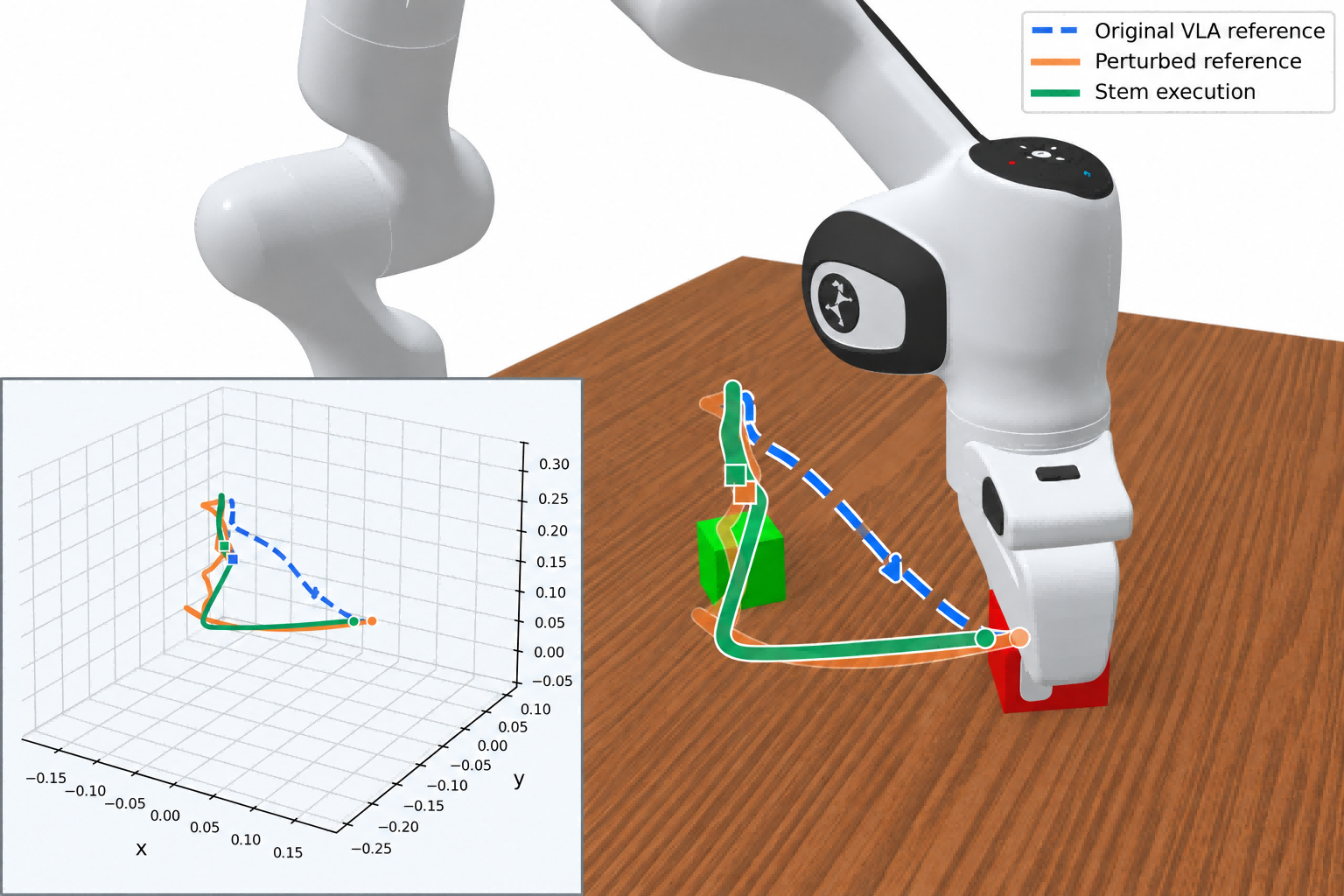}
            \par\vspace{0.1em}
            {\small (a) Trajectory visualization.}
        \end{minipage}
        \hfill
        \begin{minipage}[c]{0.49\linewidth}
            \centering
            \scriptsize
            \renewcommand{\arraystretch}{1.35}
            \setlength{\tabcolsep}{2pt}
            \resizebox{\linewidth}{!}{%
                \begin{tabular}{lccc}
                    \toprule
                    \textbf{Dir.} &
                    $D_{\mathrm{mod}}^{\mathrm{norm}}\downarrow$ &
                    $D_{\mathrm{ori}}^{\mathrm{norm}}\downarrow$ &
                    SUCC$\uparrow$ \\
                    \midrule
                    Left  & $0.218\!\pm\!0.189$ & $0.954\!\pm\!1.165$ & 60.0\% \\
                    Right & $0.269\!\pm\!0.292$ & $0.920\!\pm\!0.913$ & 63.3\% \\
                    Up    & $0.203\!\pm\!0.258$ & $0.808\!\pm\!1.248$ & 70.0\% \\
                    \bottomrule
                \end{tabular}%
            }
            \par\vspace{0.4em}
            {\small (b) Quantitative results.}
        \end{minipage}%
    }

    \vspace{-0.3em}
    \caption{
        Reference-conditioned execution on StackCube:
        trajectory visualization and quantitative results over
        30 paired trials per direction.
    }
    \label{fig:reference_conditioning}
\end{figure}

\subsubsection{Analysis and Ablations}
\label{sec:analysis_ablation}


Beyond overall safety--task performance, we are particularly interested in how the Stem policy encodes and conditions on the motion information contained in VLA action chunks.
Specifically, we examine how VLA-provided motion references shape the Stem policy's capabilities during training and continue to guide its execution at inference time.
We further investigate the respective roles of trajectory-level guidance and teacher--student distillation in supporting this learning process.

\begin{table}[t]
\centering
\caption{VLA reference ablation on PickCube.
Results are reported for student policies.}
\label{tab:ablation_reference}

\renewcommand{\arraystretch}{0.9}
\setlength{\tabcolsep}{5pt}
\scriptsize

\resizebox{\columnwidth}{!}{%
\begin{tabular}{
>{\centering\arraybackslash}p{3.2cm}
*{6}{c}
}
\toprule
\multirow{2}{*}{\textbf{Method}} &
\multicolumn{2}{c}{\textbf{Normal}} &
\multicolumn{2}{c}{\textbf{Static}} &
\multicolumn{2}{c}{\textbf{Dynamic}} \\
\cmidrule(lr){2-3}
\cmidrule(lr){4-5}
\cmidrule(lr){6-7}
& SUCC$\uparrow$ & COLL$\downarrow$
& SUCC$\uparrow$ & COLL$\downarrow$
& SUCC$\uparrow$ & COLL$\downarrow$ \\
\midrule

\sysname w/o reference
& 0.0000 & --
& 0.0000 & 0.1300
& 0.0000 & 0.0156 \\

\sysname
& \textbf{0.7325} & --
& \textbf{0.5219} & 0.2316
& \textbf{0.4153} & 0.3122 \\

\bottomrule
\end{tabular}%
}
\end{table}

\begin{figure*}[t]
    \centering

    \makebox[\textwidth][c]{%
        \begin{minipage}[t]{0.36\textwidth}
            \centering
            \includegraphics[width=\linewidth]{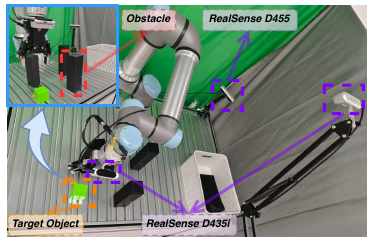}\\[-0.3em]
            {\small (a) Real-world examples.}
        \end{minipage}
        \hspace{0.015\textwidth}
        \begin{minipage}[t]{0.30\textwidth}
            \centering
            \includegraphics[width=\linewidth]{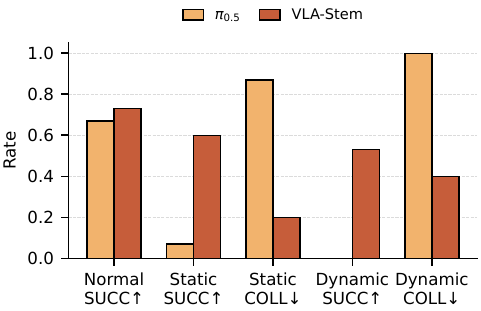}\\[-0.3em]
            {\small (b) Object Stowing.}
        \end{minipage}
        \hspace{0.015\textwidth}
        \begin{minipage}[t]{0.30\textwidth}
            \centering
            \includegraphics[width=\linewidth]{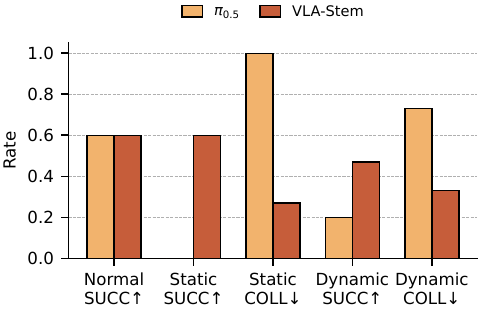}\\[-0.3em]
            {\small (c) Stacking.}
        \end{minipage}%
    }

    \vspace{-0.4em}
    \caption{Real-world evaluation of VLA-Stem.
    (a) Real-world experimental examples.
    (b--c) Quantitative results on Object Stowing and Stacking, respectively.}
    \label{fig:real_world}
\end{figure*}

\textbf{Reference-conditioned execution.}
We first examine whether the Stem policy relies on VLA-provided motion references to develop its task-execution capability.
On PickCube, we remove both the VLA reference inputs and reference-based guidance rewards from teacher--student training, while keeping the same task settings, network architecture, and remaining task and collision rewards unchanged.
As shown in Tab.~\ref{tab:ablation_reference}, the resulting student policies achieve zero SUCC across all three scenarios despite low collision rates in the obstacle settings, while the teachers attain only a few percent success at best.
In our framework, VLA references convey task-directed motion to the student, which lacks explicit task-level observations, while the associated dense guidance directs exploration under a sparse task-reward design shared across tasks.
The poor performance after removing both signals supports their role in shaping the Stem policy's task-execution capability during training.

We next turn to the full \sysname model and examine whether its trained Stem policy remains guided by the VLA's motion reference during execution.
We design a controlled trajectory-intervention experiment to directly probe how the Stem policy responds to spatial variations of the VLA reference.
To make the spatial response clearly observable, we select a VLA action chunk with relatively large motion magnitude in StackCube and apply controlled perturbations to its trajectory.

We smoothly deform the selected VLA reference toward three spatial directions (left, right, and up), using three predefined displacement magnitudes of 15, 17.5, and 20\,cm.
For each direction, we evaluate 30 paired rollouts, with 10 pairs at each perturbation magnitude.
The execution preceding the intervention is identical within each pair; only the selected reference chunk is modified, while subsequent VLA chunks are normally regenerated from the updated observations.

For each perturbed rollout, we measure the forward-segment positional distance between the executed trajectory and the modified reference,
$D_{\mathrm{mod}}$, as well as that to the original VLA reference,
$D_{\mathrm{ori}}$.
To account for variations in VLA motion scale across episodes, both distances are normalized by the complete arc length of the corresponding original VLA reference chunk.

As shown in Fig.~\ref{fig:reference_conditioning}, Stem execution consistently follows the spatial variation introduced into the VLA reference: $D_{\mathrm{mod}}^{\mathrm{norm}}$ is on average 0.230 across left/right/up perturbations, compared with 0.894 for $D_{\mathrm{ori}}^{\mathrm{norm}}$.
Overall, 88 of 90 rollouts (97.8\%) are closer to the modified reference, showing that the VLA action chunk provides an effective spatial conditioning signal for Stem execution.

The trajectory visualization further shows that, after the local perturbation, subsequent VLA references generated from the updated robot state redirect execution toward task completion.
Together with the 60.0\%--70.0\% task success under these deliberately altered references, this demonstrates that responsive Stem execution remains coupled with subsequent VLA task progression.

\textbf{Effectiveness of trajectory-level guidance.}
To verify the effectiveness of trajectory-level guidance, we compare it with \textit{step-wise action alignment}, which penalizes the Euclidean distance between the Stem action and the corresponding VLA action at each time step, with all other settings unchanged.

\begin{table}[t]
\centering
\caption{Ablation results of Trajectory-level guidance on Object Stowing.}
\label{tab:ablation_guidance}

\renewcommand{\arraystretch}{0.6}
\setlength{\tabcolsep}{5pt}
\scriptsize  

\resizebox{\columnwidth}{!}{
\begin{tabular}{
>{\centering\arraybackslash}p{3.2cm}
*{6}{c}
}
\toprule
\multirow{2}{*}{\textbf{Method}} &
\multicolumn{2}{c}{\textbf{Normal}} &
\multicolumn{2}{c}{\textbf{Static}} &
\multicolumn{2}{c}{\textbf{Dynamic}} \\
\cmidrule(lr){2-3}
\cmidrule(lr){4-5}
\cmidrule(lr){6-7}
& SUCC$\uparrow$ & COLL$\downarrow$
& SUCC$\uparrow$ & COLL$\downarrow$
& SUCC$\uparrow$ & COLL$\downarrow$ \\
\midrule
\sysname w/ Step-wise
& 0.2888 & -- & 0.4778 & \textbf{0.0366} & 0.1453 & 0.2194 \\

\sysname
& \textbf{0.7061} & -- & \textbf{0.5784} & 0.1756 & \textbf{0.5950} & \textbf{0.1203} \\
\bottomrule
\end{tabular}}
\end{table}

Tab.~\ref{tab:ablation_guidance} shows that trajectory-level guidance substantially improves task success across all three scenarios, increasing the average success rate from 30\% to 63\% compared with step-wise action alignment.
Although step-wise alignment obtains a lower collision rate in the static scenario, its much lower success rate suggests that this does not reflect a better safety--task trade-off, but is largely associated with incomplete task execution under time-indexed strict action matching.
By replacing strict time-indexed action matching with a soft spatial constraint around the base VLA trajectory, \sysname allows flexible local motion and timing adjustments while maintaining consistency with the task-directed motion of the base policy, leading to stronger overall performance under both static constraints and dynamic obstacles.

\textbf{Necessity of teacher--student distillation.}
We compare \sysname with a variant that removes teacher--student distillation and directly trains the Stem policy using only deployable observations and the same RL rewards, while keeping all other settings unchanged.

\begin{table}[t]
\centering
\caption{Ablation results of Teacher--student distillation on Object Stowing.}
\label{tab:ablation_distillation}

\renewcommand{\arraystretch}{0.6}
\setlength{\tabcolsep}{5pt}
\scriptsize

\resizebox{\columnwidth}{!}{
\begin{tabular}{
>{\centering\arraybackslash}p{3.4cm}
*{6}{c}
}
\toprule
\multirow{2}{*}{\textbf{Method}} &
\multicolumn{2}{c}{\textbf{Normal}} &
\multicolumn{2}{c}{\textbf{Static}} &
\multicolumn{2}{c}{\textbf{Dynamic}} \\
\cmidrule(lr){2-3}
\cmidrule(lr){4-5}
\cmidrule(lr){6-7}
& SUCC$\uparrow$ & COLL$\downarrow$
& SUCC$\uparrow$ & COLL$\downarrow$
& SUCC$\uparrow$ & COLL$\downarrow$ \\
\midrule
\sysname w/o distillation
& 0.2903 & -- & 0.0525 & 0.3066 & 0.0900 & 0.4419 \\

\sysname
& \textbf{0.7061} & -- & \textbf{0.5784} & \textbf{0.1756} & \textbf{0.5950} & \textbf{0.1203} \\
\bottomrule
\end{tabular}}
\end{table}

As shown in Tab.~\ref{tab:ablation_distillation}, removing teacher--student distillation leads to substantially lower success rates and higher collision rates across all scenarios.
Compared with the variant without distillation, \sysname increases the average success rate from 14\% to 63\% and reduces the average collision rate from 37\% to 15\% in constrained scenarios.
These results indicate that privileged task observations provide more informative training signals for learning recovery behavior after obstacle-avoidance-induced deviations, which is then transferred to the deployable student policy through distillation.

\subsection{Real-World Evaluation}

We evaluate \sysname across two real-world manipulation tasks, object stowing and stacking, on a real UR7e robotic arm equipped with two RealSense D435i cameras for VLA observations.
For each task, we collect 50 teleoperated demonstrations to fine-tune the real-world VLA policy.
During execution, obstacle poses are estimated using AprilTag~\cite{krogius2019iros} with an additional RealSense D455 camera and converted into local obstacle representations.
The safety policy is trained only in simulation and directly deployed on the real robot, aided by low-dimensional inputs with smaller domain gaps than raw visual observations.

We use the same SUCC and COLL metrics as in simulation and
conduct 15 trials per task and scenario.
Fig.~\ref{fig:real_world} reports results on both object stowing and stacking.
In normal settings, \sysname retains the observed base-policy success rates.
Under static and dynamic obstacles across both tasks, \sysname reduces
collision rates from 73--100\% to 20--40\% and increases task success
from 0--20\% to 47--60\%.
These results further demonstrate the effectiveness of the simulation-trained Stem policy in real-world deployment.

\section{Conclusion}



We propose \sysname, a lightweight RL-based safety mechanism that couples frozen VLA motion references with responsive obstacle avoidance. Trajectory-level guidance allows local spatial and timing adjustments while preserving the intended motion progression, and teacher--student training supports recovery from execution deviations. Simulation and real-world experiments demonstrate improved collision-free task success under static and dynamic obstacles while retaining base-policy performance in obstacle-free settings. Direct deployment of the simulation-trained Stem policy on the real robot further demonstrates its practical potential. The current evaluation focuses on structured manipulation tasks and collision avoidance using low-dimensional obstacle representations. Future work will extend the framework to longer-horizon tasks and contact- or force-related safety objectives, while integrating robust visual perception to improve reliability under noisy obstacle observations.

\bibliographystyle{IEEEtran}
\bibliography{references}

@article{pi0.5,
  title={$\pi_{0.5}$: a Vision-Language-Action Model with Open-World Generalization},
  author={{Physical Intelligence} and Black, Kevin and Brown, Noah and Darpinian, James and Dhabalia, Karan and Driess, Danny and Esmail, Adnan and Equi, Michael and Finn, Chelsea and Fusai, Niccolo and others},
  journal={arXiv preprint arXiv:2504.16054},
  year={2025}
}

@article{kim2024openvla,
  title={{OpenVLA}: An open-source vision-language-action model},
  author={Kim, Moo Jin and Pertsch, Karl and Karamcheti, Siddharth and Xiao, Ted and Balakrishna, Ashwin and Nair, Suraj and Rafailov, Rafael and Foster, Ethan and Lam, Grace and Sanketi, Pannag and others},
  journal={arXiv preprint arXiv:2406.09246},
  year={2024}
}

@article{brohan2022rt,
  title={{RT-1}: Robotics transformer for real-world control at scale},
  author={Brohan, Anthony and Brown, Noah and Carbajal, Justice and Chebotar, Yevgen and Dabis, Joseph and Finn, Chelsea and Gopalakrishnan, Keerthana and Hausman, Karol and Herzog, Alex and Hsu, Jasmine and others},
  journal={arXiv preprint arXiv:2212.06817},
  year={2022}
}

@inproceedings{zitkovich2023rt,
  title={{RT-2}: Vision-language-action models transfer web knowledge to robotic control},
  author={Zitkovich, Brianna and Yu, Tianhe and Xu, Sichun and Xu, Peng and Xiao, Ted and Xia, Fei and Wu, Jialin and Wohlhart, Paul and Welker, Stefan and Wahid, Ayzaan and others},
  booktitle={Conference on Robot Learning},
  pages={2165--2183},
  year={2023},
  organization={PMLR}
}

@article{liu2024rdt,
  title={{RDT-1B}: a diffusion foundation model for bimanual manipulation},
  author={Liu, Songming and Wu, Lingxuan and Li, Bangguo and Tan, Hengkai and Chen, Huayu and Wang, Zhengyi and Xu, Ke and Su, Hang and Zhu, Jun},
  journal={arXiv preprint arXiv:2410.07864},
  year={2024}
}

@article{black2024pi_0,
  title={$\pi_0$: A Vision-Language-Action Flow Model for General Robot Control},
  author={Black, Kevin and Brown, Noah and Driess, Danny and Esmail, Adnan and Equi, Michael and Finn, Chelsea and Fusai, Niccolo and Groom, Lachy and Hausman, Karol and Ichter, Brian and others},
  journal={arXiv preprint arXiv:2410.24164},
  year={2024}
}

@article{amin2025pi,
  title={$\pi_{0.6}$: a {VLA} That Learns From Experience},
  author={Amin, Ali and Aniceto, Raichelle and Balakrishna, Ashwin and Black, Kevin and Conley, Ken and Connors, Grace and Darpinian, James and Dhabalia, Karan and DiCarlo, Jared and Driess, Danny and others},
  journal={arXiv preprint arXiv:2511.14759},
  year={2025}
}

@misc{lu2025vlarl,
  title={{VLA-RL}: Towards Masterful and General Robotic Manipulation with Scalable Reinforcement Learning},
  author={Guanxing Lu and Chubin Zhang and Haonan Jiang and Yuheng Zhou and Zifeng Gao and Yansong Tang and Ziwei Wang},
  year={2025},
  note={{Notion Blog}}
}

@article{lavalle2001rapidly,
  title={Rapidly-Exploring Random Trees: Progress and Prospects},
  author={LaValle, Steven M and Kuffner, James J},
  journal={Algorithmic and Computational Robotics},
  pages={303--307},
  year={2001},
  publisher={AK Peters/CRC Press}
}

@article{kavraki1996probabilistic,
  title={Probabilistic roadmaps for path planning in high-dimensional configuration spaces},
  author={Kavraki, Lydia E and Svestka, Petr and Latombe, J-C and Overmars, Mark H},
  journal={IEEE Transactions on Robotics and Automation},
  volume={12},
  number={4},
  pages={566--580},
  year={1996},
  publisher={IEEE}
}

@inproceedings{ratliff2009chomp,
  title={{CHOMP}: Gradient optimization techniques for efficient motion planning},
  author={Ratliff, Nathan and Zucker, Matt and Bagnell, J Andrew and Srinivasa, Siddhartha},
  booktitle={2009 IEEE International Conference on Robotics and Automation},
  pages={489--494},
  year={2009},
  organization={IEEE}
}

@article{TrajOpt,
  title={Motion planning with sequential convex optimization and convex collision checking},
  author={Schulman, John and Duan, Yan and Ho, Jonathan and Lee, Alex and Awwal, Ibrahim and Bradlow, Henry and Pan, Jia and Patil, Sachin and Goldberg, Ken and Abbeel, Pieter},
  journal={The International Journal of Robotics Research},
  volume={33},
  number={9},
  pages={1251--1270},
  year={2014},
  publisher={Sage Publications Sage UK: London, England}
}

@inproceedings{carvalho2023motionplanningdiffusion,
  title={Motion planning diffusion: Learning and planning of robot motions with diffusion models},
  author={Carvalho, Joao and Le, An T and Baierl, Mark and Koert, Dorothea and Peters, Jan},
  booktitle={2023 IEEE/RSJ International Conference on Intelligent Robots and Systems (IROS)},
  pages={1916--1923},
  year={2023},
  organization={IEEE}
}

@inproceedings{kalakrishnan2011stomp,
  title={{STOMP}: Stochastic trajectory optimization for motion planning},
  author={Kalakrishnan, Mrinal and Chitta, Sachin and Theodorou, Evangelos and Pastor, Peter and Schaal, Stefan},
  booktitle={2011 IEEE International Conference on Robotics and Automation},
  pages={4569--4574},
  year={2011},
  organization={IEEE}
}

@article{ying2025urplanner,
  title={{URPlanner}: A Universal Paradigm For Collision-Free Robotic Motion Planning Based on Deep Reinforcement Learning},
  author={Ying, Fengkang and Zhang, Hanwen and Wang, Haozhe and Huang, Huishi and Ang Jr, Marcelo H},
  journal={arXiv preprint arXiv:2505.20175},
  year={2025}
}

@article{yang2025deep,
  title={Deep reactive policy: Learning reactive manipulator motion planning for dynamic environments},
  author={Yang, Jiahui and Liu, Jason Jingzhou and Li, Yulong and Khaky, Youssef and Shaw, Kenneth and Pathak, Deepak},
  journal={arXiv preprint arXiv:2509.06953},
  year={2025}
}

@inproceedings{saferl1,
  title={End-to-end safe reinforcement learning through barrier functions for safety-critical continuous control tasks},
  author={Cheng, Richard and Orosz, G{\'a}bor and Murray, Richard M and Burdick, Joel W},
  booktitle={Proceedings of the AAAI Conference on Artificial Intelligence},
  volume={33},
  number={01},
  pages={3387--3395},
  year={2019}
}

@inproceedings{saferl2,
  title={Predictive control barrier functions for online safety critical control},
  author={Breeden, Joseph and Panagou, Dimitra},
  booktitle={2022 IEEE 61st Conference on Decision and Control (CDC)},
  pages={924--931},
  year={2022},
  organization={IEEE}
}

@inproceedings{jung2025rail,
  title={{RAIL}: Reachability-aided imitation learning for safe policy execution},
  author={Jung, Wonsuhk and Anthony, Dennis and Mishra, Utkarsh A and Arachchige, Nadun Ranawaka and Bronars, Matthew and Xu, Danfei and Kousik, Shreyas},
  booktitle={2025 IEEE International Conference on Robotics and Automation (ICRA)},
  pages={3582--3589},
  year={2025},
  organization={IEEE}
}

@article{deng2025survey,
  title={A Survey on Reinforcement Learning of {Vision-Language-Action} Models for Robotic Manipulation},
  author={Deng, Haoyuan and Wu, Zhenyu and Liu, Haichao and Guo, Wenkai and Xue, Yuquan and Shan, Ziyu and Zhang, Chuanrui and Jia, Bofang and Ling, Yuan and Lu, Guanxing and others},
  journal={Authorea Preprints},
  year={2025},
  publisher={Authorea}
}

@article{guo2025vlareasoner,
  title={{VLA-Reasoner}: Empowering vision-language-action models with reasoning via online Monte Carlo tree search},
  author={Guo, Wenkai and Lu, Guanxing and Deng, Haoyuan and Wu, Zhenyu and Tang, Yansong and Wang, Ziwei},
  journal={arXiv preprint arXiv:2509.22643},
  year={2025}
}

@article{nakamoto2024steering,
  title={Steering your generalists: Improving robotic foundation models via value guidance},
  author={Nakamoto, Mitsuhiko and Mees, Oier and Kumar, Aviral and Levine, Sergey},
  journal={arXiv preprint arXiv:2410.13816},
  year={2024}
}

@article{ze2025twist,
  title={Twist: Teleoperated whole-body imitation system},
  author={Ze, Yanjie and Chen, Zixuan and Ara{\'u}jo, Joao Pedro and Cao, Zi-ang and Peng, Xue Bin and Wu, Jiajun and Liu, C Karen},
  journal={arXiv preprint arXiv:2505.02833},
  year={2025}
}

@article{schulman2017proximal,
  title={Proximal policy optimization algorithms},
  author={Schulman, John and Wolski, Filip and Dhariwal, Prafulla and Radford, Alec and Klimov, Oleg},
  journal={arXiv preprint arXiv:1707.06347},
  year={2017}
}

@inproceedings{bengio2009curriculum,
  title={Curriculum learning},
  author={Bengio, Yoshua and Louradour, J{\'e}r{\^o}me and Collobert, Ronan and Weston, Jason},
  booktitle={Proceedings of the 26th Annual International Conference on Machine Learning},
  pages={41--48},
  year={2009}
}

@article{taomaniskill3,
  title={{ManiSkill3}: {GPU} Parallelized Robotics Simulation and Rendering for Generalizable Embodied {AI}},
  author={Stone Tao and Fanbo Xiang and Arth Shukla and Yuzhe Qin and Xander Hinrichsen and Xiaodi Yuan and Chen Bao and Xinsong Lin and Yulin Liu and Chan, Tse-kai and Yuan Gao and Xuanlin Li and Tongzhou Mu and Nan Xiao and Arnav Gurha and Viswesh Nagaswamy Rajesh and Yong Woo Choi and Yen-Ru Chen and Zhiao Huang and Roberto Calandra and Rui Chen and Shan Luo and Hao Su},
  journal={Robotics: Science and Systems},
  year={2025}
}

@inproceedings{krogius2019iros,
    TITLE      = {Flexible Layouts for Fiducial Tags},
    AUTHOR     = {Maximilian Krogius and Acshi Haggenmiller and Edwin Olson},
    BOOKTITLE  = {Proceedings of the {IEEE/RSJ} International Conference on Intelligent
                 Robots and Systems {(IROS)}},
    YEAR       = {2019},
    MONTH      = {October},
}

@article{hu2025vlsa,
  title={{VLSA}: {Vision-Language-Action} models with plug-and-play safety constraint layer},
  author={Hu, Songqiao and Liu, Zeyi and Liu, Shuang and Cen, Jun and Meng, Zihan and Wang, Shihefeng and Li, Xiang and He, Xiao},
  journal={arXiv preprint arXiv:2512.11891},
  year={2025}
}

@article{zhang2026safevla,
  title={{SafeVLA}: Towards safety alignment of vision-language-action model via constrained learning},
  author={Zhang, Borong and Zhang, Yuhao and Ji, Jiaming and Lei, Yingshan and Dai, Juntao and Chen, Yuanpei and Yang, Yaodong},
  journal={Advances in Neural Information Processing Systems},
  volume={38},
  pages={153335--153373},
  year={2026}
}

@inproceedings{yu2024ldp,
  title={Ldp: A local diffusion planner for efficient robot navigation and collision avoidance},
  author={Yu, Wenhao and Peng, Jie and Yang, Huanyu and Zhang, Junrui and Duan, Yifan and Ji, Jianmin and Zhang, Yanyong},
  booktitle={2024 IEEE/RSJ international conference on intelligent robots and systems (IROS)},
  pages={5466--5472},
  year={2024},
  organization={IEEE}
}

@inproceedings{yu2024pathrl,
  title={Pathrl: An end-to-end path generation method for collision avoidance via deep reinforcement learning},
  author={Yu, Wenhao and Peng, Jie and Qiu, Quecheng and Wang, Hanyu and Zhang, Lu and Ji, Jianmin},
  booktitle={2024 IEEE International Conference on Robotics and Automation (ICRA)},
  pages={9278--9284},
  year={2024},
  organization={IEEE}
}

@article{xiong2026cac,
  title={CAC-VLA: Context-Gated Action Conditioning for Vision-Language-Action Models},
  author={Xiong, Yifu and Yu, Wenhao and Lin, Jiaxuan and Zou, Bojun and Li, Jiahao and Zhang, Lu and Zhang, Yanyong and Ji, Jianmin},
  journal={arXiv preprint arXiv:2607.04816},
  year={2026}
}

@article{geng2026trajwam,
  title={TrajWAM: An Efficient Trajectory-Guided World Action Model for Dynamic Robotic Manipulation},
  author={Geng, Haohan and Zhang, Sha and Sun, Weishuo and Ye, Guanting and Yu, Wenhao and Xie, Yuang and Zeng, Qicheng and Ye, Peng and Deng, Jiajun and Ji, Jianmin},
  journal={arXiv preprint arXiv:2610.25763},
  year={2026}
}

@article{ye2026c,
  title={{C$^{2}$RoPE}: Causal Continuous Rotary Positional Encoding for 3D Large Multimodal-Models Reasoning},
  author={Ye, Guanting and Zhao, Qiyan and Yu, Wenhao and Zhang, Xiaofeng and Ji, Jianmin and Zhang, Yanyong and Yuen, Ka-Veng},
  journal={arXiv preprint arXiv:2602.10551},
  year={2026}
}

@inproceedings{yip2026sope,
  title={Sope: Spherical coordinate-based positional embedding for enhancing spatial perception of 3d lvlms},
  author={Yip, Koonting and Zhao, Qiyan and Yu, Wenhao and Yuen, Liangyu and Li, Mingkai and Zhang, Xiaofeng and Ji, Jianmin and Zhang, Yanyong and Jiang, Qing and Yuen, Ka-Veng},
  booktitle={Proceedings of the IEEE/CVF Conference on Computer Vision and Pattern Recognition},
  pages={33714--33726},
  year={2026}
}

\clearpage
\appendix

\subsection{Implementation and Execution Details}
\label{app:implementation}

\subsubsection{VLA-Conditioned Execution}
\label{app:execution}

VLA-Stem couples a frozen vision--language--action (VLA) policy with a separately trained Stem policy. At the beginning $t_c$ of each execution interval $c$, the base VLA generates an action chunk of length $h=8$. As described in Sec.~III-A, the proposed actions are converted into future end-effector poses and expressed relative to the end-effector pose at the beginning of the interval:
\begin{equation}
    \Delta T_{c,k}
    = (T^w_{t_c})^{-1} T^w_{t_c+k}, \qquad
    \tau_c^{\mathrm{rel}} = \{\Delta T_{c,k}\}_{k=1}^{h}.
\end{equation}

The resulting trajectory serves as a motion reference rather than being executed directly. At each control step $t$ within the interval, the Stem policy generates an executable action conditioned on robot proprioception $x_t$, local obstacle information $o_t^{\mathrm{obs}}$, and the active reference $\tau_c^{\mathrm{rel}}$:
\begin{equation}
    a_t^{\mathrm{exec}}
    = \pi_{\mathrm{stem}}(x_t,o_t^{\mathrm{obs}},\tau_c^{\mathrm{rel}}),
    \quad t_c\leq t<t_c+h.
\end{equation}

The deployable Stem policy uses three low-dimensional inputs. The proprioceptive input is a 16-dimensional vector consisting of nine joint-position values and a seven-dimensional end-effector pose. The VLA reference is represented as an $8\times8$ trajectory feature sequence. The obstacle input is a 33-dimensional vector supporting up to three obstacles, with 11 features per obstacle slot. These features include the obstacle geometry described in Sec.~III-A. The inputs are processed without raw visual observations.

The Stem policy outputs a seven-dimensional normalized action under the end-effector delta-pose control mode. The action is bounded to $[-1,1]$ through a hyperbolic tangent function. The complete executable action is generated by the Stem policy, while the VLA action chunk provides the motion reference for conditioning. After each execution interval, the frozen VLA generates a new reference based on the updated observation.

\subsubsection{Policy Architecture}
\label{app:architecture}

The deployable Stem policy adopts an MLP-based actor--critic architecture with three independent feature encoders. The VLA reference trajectory is flattened and processed by a three-layer MLP with hidden dimensions $(256,256,256)$. The proprioceptive and obstacle inputs are encoded by two-layer MLPs with dimensions $(256,256)$ and $(64,64)$, respectively. ReLU activations are used between layers.

The encoded features are concatenated into a 576-dimensional representation shared by the actor and critic. Both heads contain three hidden layers of dimensions $(1024,1024,512)$ with ReLU activations. The actor parameterizes a seven-dimensional diagonal Gaussian action distribution with a learned, state-independent log standard deviation. Actions are passed through a tanh transformation, while deterministic inference uses the tanh-transformed action mean. The critic outputs a scalar state-value estimate.

The deployable Stem policy contains approximately 4.56 million trainable parameters. During teacher--student training, the privileged teacher additionally receives task-related state information, resulting in a 23-dimensional state input, whereas the deployed student uses only the 16-dimensional proprioceptive input.

\subsection{RL Training and Reward Details}
\label{app:reward}

\subsubsection{Reward Components}
\label{app:reward_components}

The reward defined in Sec.~III-D is implemented by separating the trajectory-level guidance into trajectory-following and progress terms:
\begin{equation}
\begin{aligned}
    r_t
    &= w_{\mathrm{task}}r_t^{\mathrm{task}}
    + w_{\mathrm{coll}}r_t^{\mathrm{coll}} \\
    &\quad + w_{\mathrm{follow}}r_t^{\mathrm{follow}}
    + w_{\mathrm{step}}r_t^{\mathrm{step}}.
\end{aligned}
\label{eq:app_reward}
\end{equation}
The last two terms jointly correspond to the guidance contribution $w_{\mathrm{guidance}}r_t^{\mathrm{guidance}}$ in the main formulation.

The task reward consists of a sparse grasping bonus and a terminal success reward:
\begin{equation}
    r_t^{\mathrm{task}}
    = 0.4r_t^{\mathrm{grasp}} + r_t^{\mathrm{succ}}.
\end{equation}
Here, $r_t^{\mathrm{grasp}}=1$ when the target object is first successfully grasped, and $r_t^{\mathrm{succ}}=1$ upon task completion; otherwise, both terms are zero. Each event is rewarded only once per episode.

The collision penalty is defined as
\begin{equation}
    r_t^{\mathrm{coll}} = -r_t^{\mathrm{viol}},
\end{equation}
where $r_t^{\mathrm{viol}}=1$ if a collision or spatial-constraint violation occurs at step $t$, and $0$ otherwise.

The trajectory-following reward encourages spatial consistency with the VLA reference, while the auxiliary progress reward encourages forward advancement along the reference trajectory during early curriculum stages. Their detailed computations are provided in App.~\ref{app:guidance_reward}.

The reward weights used in each curriculum stage are summarized in Tab.~\ref{tab:reward_weights}.

\begin{table}[htbp]
\centering
\caption{Reward weights used in the three curriculum stages.}
\label{tab:reward_weights}
\renewcommand{\arraystretch}{1.08}
\setlength{\tabcolsep}{5pt}
\footnotesize
\begin{tabular}{@{}lcccc@{}}
\toprule
\textbf{Stage} &
$w_{\mathrm{task}}$ &
$w_{\mathrm{coll}}$ &
$w_{\mathrm{follow}}$ &
$w_{\mathrm{step}}$ \\
\midrule
Following & 0.6 & 0.0 & 0.4 & 0.2 \\
Static    & 1.2 & 1.0 & 0.4 & 0.1 \\
Dynamic   & 1.2 & 1.0 & 0.6 & 0.0 \\
\bottomrule
\end{tabular}
\end{table}

\subsubsection{PPO Optimization and Training Budget}
\label{app:ppo_training}

The teacher and student policies are optimized using PPO with GPU-parallel simulation and the Adam optimizer. The stage-wise training configurations and simulation budgets are summarized in Tab.~\ref{tab:ppo_hyperparams}. The reported environment transitions include both teacher and student training.

\begin{table}[htbp]
\centering
\caption{Stage-wise PPO training configurations and simulation budgets. Transition counts are approximate and include both teacher and student training.}
\label{tab:ppo_hyperparams}
\renewcommand{\arraystretch}{1.08}
\setlength{\tabcolsep}{3pt}
\footnotesize
\begin{tabular}{@{}lccc@{}}
\toprule
\textbf{Parameter} &
\textbf{Following} &
\textbf{Static} &
\textbf{Dynamic} \\
\midrule
Parallel environments & 256 & 192 & 192 \\
Learning rate & $2\!\times\!10^{-4}$ & $1.5\!\times\!10^{-4}$ & $1.5\!\times\!10^{-4}$ \\
PPO update epochs & 8 & 12 & 12 \\
Minibatch size & 3,200 & 19,200 & 19,200 \\
Rollout horizon & 100 & 100 & 100 \\
Discount factor $\gamma$ & 0.8 & 0.8 & 0.8 \\
Transitions / seed & $\sim$38.0M & $\sim$23.7M & $\sim$20.6M \\
\bottomrule
\end{tabular}
\end{table}

The student PPO implementation uses generalized advantage estimation (GAE) with $\lambda_{\mathrm{GAE}}=0.9$, a clipping coefficient of $0.2$, a value-loss coefficient of $0.5$, and an entropy coefficient of $0$. Gradients are clipped to a maximum norm of $0.5$, and the Adam optimizer uses $\epsilon=10^{-5}$.


\subsubsection{Teacher--Student Distillation}
\label{app:teacher_student}

The privileged teacher is first optimized with PPO under the curriculum described in Sec.~III-D. It receives the deployable observation $o_t$ together with privileged task-level information $o_{\mathrm{extra},t}$. After training, the teacher is frozen and used to guide the optimization of the deployable student.

During student training, both deployable observations and their corresponding privileged observations are collected in the rollout buffer. The frozen teacher is evaluated on these observations during PPO minibatch updates, providing action distributions for distillation. The student is jointly optimized using the PPO objective and the distillation loss:
\begin{equation}
    L(\pi_{\mathrm{stu}})
    =
    L_{\mathrm{PPO}}(\pi_{\mathrm{stu}};\mathcal{D})
    +
    \lambda L_{\mathrm{distill}}.
    \label{eq:app_student_loss}
\end{equation}
The distillation term is defined as
\begin{equation}
\begin{aligned}
    L_{\mathrm{distill}}
    = \mathbb{E}_{\mathcal{D}}\bigl[
    &D_{\mathrm{KL}}\bigl(
    \pi_{\mathrm{stu}}(\cdot\mid o)
    \,\Vert\\
    &\qquad
    \pi_{\mathrm{tea}}(\cdot\mid o,o_{\mathrm{extra}})
    \bigr)\bigr].
\end{aligned}
\end{equation}
Here, $\mathcal{D}$ denotes the student rollout buffer, and $L_{\mathrm{PPO}}$ includes the clipped policy objective, value-function regression, and entropy regularization. The distillation coefficient is set to $\lambda=0.1$.

The privileged inputs are used only for teacher supervision and do not modify the reward function. After distillation, only the student policy is deployed, without requiring privileged task information.

\subsection{Computation of Trajectory-Level Guidance}
\label{app:guidance_reward}

\subsubsection{Pose Deviation to Reference Segments}

Consider the active world-frame reference chunk for interval $c$,
\begin{equation}
    \tau_c^{\mathrm{ref}}
    =\{(\mathbf{p}_i^{\mathrm{ref}},q_i^{\mathrm{ref}})\}_{i=1}^{h}.
\end{equation}
At a control step $t$ within this interval, we compute the deviation between the current end-effector pose $(\mathbf{p}_t,q_t)$ and each segment of $\tau_c^{\mathrm{ref}}$. Positional waypoints define piecewise linear segments, and their orientations are interpolated using normalized quaternion interpolation.

For segment $i\in\{1,\ldots,h-1\}$, let $A_i=\mathbf{p}_i^{\mathrm{ref}}$ and $B_i=\mathbf{p}_{i+1}^{\mathrm{ref}}$. We project $\mathbf{p}_t$ onto the segment using
\begin{equation}
\begin{aligned}
    \lambda_i
    &=\operatorname{clip}\!\left(
        \frac{(\mathbf{p}_t-A_i)^\top(B_i-A_i)}
        {\|B_i-A_i\|^2+\epsilon},0,1
      \right),\\
    \hat{\mathbf{p}}_i
    &=A_i+\lambda_i(B_i-A_i),
\end{aligned}
\end{equation}
where $\epsilon>0$ is a numerical stability constant. The positional deviation is
\begin{equation}
    d_i^{\mathrm{pos}}
    =\|\mathbf{p}_t-\hat{\mathbf{p}}_i\|.
\end{equation}

To compute the rotational deviation at the same projected point, we first account for the double-cover property of quaternions by adjusting the sign of the segment endpoint:
\begin{equation}
\begin{aligned}
    s_i & =
    \begin{cases}
        1, & \langle q_i^{\mathrm{ref}},q_{i+1}^{\mathrm{ref}}\rangle\geq 0,\\
       -1, & \langle q_i^{\mathrm{ref}},q_{i+1}^{\mathrm{ref}}\rangle<0,
    \end{cases}\\
    \tilde q_{i+1}^{\mathrm{ref}} &=s_i q_{i+1}^{\mathrm{ref}}.
\end{aligned}
\end{equation}
The interpolated orientation is obtained by normalizing the linear interpolation:
\begin{equation}
\begin{aligned}
    u_i &=(1-\lambda_i)q_i^{\mathrm{ref}}
        +\lambda_i\tilde q_{i+1}^{\mathrm{ref}},\\
    \hat q_i &=\frac{u_i}{\|u_i\|}.
\end{aligned}
\end{equation}
The geodesic rotational deviation between unit quaternions is
\begin{equation}
    d_i^{\mathrm{rot}}
    =2\arccos\!\left(
       \operatorname{clip}\bigl(|\langle q_t,\hat q_i\rangle|,0,1\bigr)
    \right).
\end{equation}
The segment-level pose deviation combines the positional and rotational components:
\begin{equation}
    d_i=d_i^{\mathrm{pos}}+\alpha d_i^{\mathrm{rot}},
    \qquad \alpha=\frac{4}{7}.
\end{equation}
In implementation, the corresponding weighted deviation is
$\tilde d_i=0.7d_i=0.7d_i^{\mathrm{pos}}+
0.4d_i^{\mathrm{rot}}$.

\subsubsection{Forward Corridor and Progress Reward}

Reference matching is performed independently for each \emph{active VLA action chunk}. At the start $t_c$ of every execution interval, the progress index is reset to $m_{t_c-1}=1$. Within the interval, the index is constrained to be non-decreasing. The candidate segment set includes the current segment and all subsequent segments:
\begin{equation}
    \mathcal{I}_t^+
    =\{m_{t-1},\ldots,
    \min(m_{t-1}+K,h-1)\}.
\end{equation}
We set $K=h-2=6$ for the reference horizon$h=8$, allowing matching against all remainingsegments without an additional look-ahead restriction.
The matched segment and minimum pose deviation are
\begin{equation}
\begin{aligned}
    i_t^* &=\arg\min_{i\in\mathcal{I}_t^+}d_i,\\
    d_t^* &=d_{i_t^*},\\
    m_t &=\max(m_{t-1},i_t^*).
\end{aligned}
\end{equation}
This matching rule prevents the selected reference segment from moving backward within a chunk, while allowing the robot to remain near the current segment when waiting or making a local detour. The auxiliary step-progress reward used in the early curriculum stages is the normalized index advancement:
\begin{equation}
    r_t^{\mathrm{step}}
    =\frac{\max(0,m_t-m_{t-1})}{h-1}.
\end{equation}
The trajectory-following reward is computed as
\begin{equation}
    r_t^{\mathrm{follow}}
    =\exp(-10\tilde d_t^*)
    =\exp(-7d_t^*),
\end{equation}
The trajectory-following and progress terms enter Eq.~\eqref{eq:app_reward} with their respective curriculum weights.

\subsection{Baseline Implementation Details}
\label{app:baselines}

\subsubsection{Base VLA Policies}

We report the raw $\pi_0$ and $\pi_{0.5}$ policies as unmodified baselines. For each simulation task, the corresponding base VLA is fine-tuned using 100 demonstrations for 30,000 optimization steps; its checkpoint is then held fixed when comparing runtime safety mechanisms and training Stem. This ensures that Stem and the safety-layer baselines are evaluated with the same task-level VLA policy.

\subsubsection{VLA Fine-Tuning Baseline}

The VLA-finetune baseline trains separate policies using planner-generated obstacle-avoidance demonstrations. We generate these trajectories using RRT-based planning; dynamic scenarios additionally employ a waiting primitive. We vary the number of obstacle-avoidance demonstrations and report the best-performing configuration in the main comparison. The evaluated performance peaks at 1,500 demonstrations, while increasing the budget to 2,000 results in a decline, as discussed in the main paper.

\subsubsection{VLSA Safety-Layer Baseline}
\label{app:vlsa}

\paragraph{Geometric CBF Formulation}
We approximate the robot and obstacles using oriented ellipsoids. VLSA-EE models only the end-effector ellipsoid, while VLSA-ARM additionally includes five arm links (\texttt{panda\_link3}--\texttt{panda\_link7}), resulting in six robot ellipsoids.

For each robot--obstacle ellipsoid pair $(i,j)$, we adopt a rotating supporting-hyperplane barrier. Let $\mathbf{p}_i$ and $\mathbf{p}_j$ denote the ellipsoid centers, $\bar{\mathbf{Q}}_i$ and $\bar{\mathbf{Q}}_j$ their world-frame shape matrices, and $\mathbf{z}_{ij}$ a virtual separation direction on the unit sphere. The barrier is defined as
\begin{equation}
\begin{aligned}
h_{ij}
&=
\frac{
(\mathbf{p}_j-\mathbf{p}_i)^\top
\bar{\mathbf{Q}}_i^{-1}\mathbf{z}_{ij}
}{
\|\bar{\mathbf{Q}}_i^{-1}\mathbf{z}_{ij}\|
}\\
&\quad-
\frac{
\|\bar{\mathbf{Q}}_j\bar{\mathbf{Q}}_i^{-1}
\mathbf{z}_{ij}\|+1
}{
\|\bar{\mathbf{Q}}_i^{-1}\mathbf{z}_{ij}\|
}.
\end{aligned}
\end{equation}
The condition $h_{ij}\geq0$ represents geometric separation along the selected virtual direction. Each robot--obstacle pair introduces a linear CBF constraint:
\begin{equation}
\begin{aligned}
&\mathbf{a}_{v,ij}^{\top}\mathbf{v}_i
+\mathbf{a}_{\omega,ij}^{\top}\boldsymbol{\omega}_i\\
&\quad+
\mathbf{a}_{z,ij}^{\top}\mathbf{u}_{z,ij}
+\alpha h_{ij}\geq0,
\end{aligned}
\end{equation}
where $\mathbf{v}_i$ and $\boldsymbol{\omega}_i$ are the linear and angular velocities of the corresponding robot ellipsoid, and $\mathbf{u}_{z,ij}$ controls the virtual separation direction. For VLSA-ARM, link poses and Jacobians are computed from the current robot configuration to construct the constraints for all modeled links. Box-shaped obstacles are approximated by ellipsoids with matching centers and orientations, with geometric scales included in parameter tuning.

\paragraph{QP Optimization}
The safety layer solves a weighted quadratic program to minimize deviations from the nominal VLA action:
\begin{equation}
\begin{aligned}
\min_{\mathbf{u},\{\mathbf{u}_{z,ij}\}}\quad
&\frac{w_m}{2}
\|\mathbf{u}-\mathbf{u}_{\mathrm{nom}}\|_2^2\\
&+\frac{w_z}{2}
\sum_{i,j}
\|\mathbf{u}_{z,ij}
-\mathbf{u}_{z,ij}^{\mathrm{nom}}\|_2^2,
\end{aligned}
\end{equation}
subject to the robot--obstacle CBF constraints. Here, $\mathbf{u}\in\mathbb{R}^6$ denotes the end-effector motion command. The gripper command is directly retained from the base VLA.

We use batched PyTorch-based QP solvers. VLSA-EE employs active-set enumeration for at most three constraints, while VLSA-ARM uses batched Hildreth dual projection to handle up to 18 constraints. Infeasible QPs are handled using a weighted half-space projection fallback. The implementation also retains the action-space and QP-space scaling transformations of the original VLSA formulation.

\paragraph{Parameter Tuning}
For each task and scenario, we tune the CBF gain, geometric scales, and optimization weights on validation episodes. The configuration achieving the highest collision-free task success (SUCC) is selected for the corresponding setting.

\paragraph{Dynamic Obstacle Handling}
In dynamic scenarios, the obstacle geometry is updated online using the current ground-truth obstacle observations. Each obstacle is represented by its center, full dimensions, orientation, and presence indicator, forming an 11-dimensional feature vector, with up to three obstacle slots.

At each control step, VLSA reconstructs the obstacle ellipsoids and updates the corresponding CBF constraints using their current poses. The implementation does not estimate obstacle velocities or predict future obstacle trajectories. It therefore performs reactive safety correction based on instantaneous geometric information rather than predictive dynamic-obstacle avoidance.

\end{document}